\documentclass[letterpaper, 10 pt, conference]{ieeeconf}  

\IEEEoverridecommandlockouts                              

\usepackage{algorithm}
\usepackage{algpseudocode}
\usepackage{amsmath}
\usepackage{amssymb}
\usepackage{color}
\usepackage{graphicx}
\usepackage{url}
\usepackage{cite}

\title{\Large \bf
BaRC: Backward Reachability Curriculum for Robotic Reinforcement Learning
}

\author{Boris Ivanovic$^1$, James Harrison$^2$, Apoorva Sharma$^1$, Mo Chen$^3$, Marco Pavone$^1$
\thanks{$^{1}$Department of Aeronautics and Astronautics, Stanford University, Stanford,
CA, USA 94305. 
        {\tt\small \{borisi,apoorva,pavone\}@stanford.edu}}%
\thanks{$^{2}$Department of Mechanical Engineering, Stanford University, Stanford,
CA, USA 94305. 
        {\tt\small jharrison@stanford.edu}}%
\thanks{$^{3}$School of Computing Science, Simon Fraser University, Burnaby, BC, Canada V5A 1S6. 
        {\tt\small mochen@cs.sfu.ca}}
}

\newcommand{\R}{\mathbb R}

\newcommand{\dyn}{f}
\newcommand{\state}{s}
\newcommand{\ctrl}{a}

\newcommand{\cset}{\mathcal A}

\newcommand{\cstate}{\hat\state}
\newcommand{\cdyn}{\hat\dyn}
\newcommand{\cctrl}{\hat\ctrl}
\newcommand{\ccset}{\hat \cset}

\newcommand{\scmap}{\phi}

\newcommand{\expA}{Car Model}
\newcommand{\expB}{Planar Quadrotor Model}

\newcommand{\ham}{H}
\newcommand{\valfunc}{V}
\newcommand{\fc}{\valfunc_0}
\newcommand{\costate}{\lambda}

\newcommand{\maxbrs}{\mathcal R}

\newcommand{\targetset}{\mathcal F}

\newcommand{\algName}{BaRC}
\newcommand{\algNameLong}{Backward Reachable Curriculum}

\newtheorem{defn}{Definition}

\renewcommand{\baselinestretch}{0.98}

\begin{document}

\maketitle
\thispagestyle{empty}
\pagestyle{empty}

\begin{abstract}
Model-free Reinforcement Learning (RL) offers an attractive approach to learn control policies for high-dimensional systems, but its relatively poor sample complexity often necessitates training in simulated environments. 
Even in simulation, goal-directed tasks whose natural reward function is sparse 
remain intractable for state-of-the-art model-free algorithms for continuous control. 
The bottleneck in these tasks is the prohibitive amount of exploration required to obtain a learning signal from the initial state of the system.
In this work, we leverage physical priors in the form of an approximate system dynamics model to design a curriculum for a model-free policy optimization algorithm.
Our Backward Reachability Curriculum (\algName{}) begins policy training from states that require a small number of actions to accomplish the task, and expands the initial state distribution backwards in a dynamically-consistent manner once the policy optimization algorithm demonstrates sufficient performance.
\algName{} is general, in that it can accelerate training of any model-free RL algorithm on a broad class of goal-directed continuous control MDPs. 
Its curriculum strategy is physically intuitive, easy-to-tune, and allows incorporating physical priors to accelerate training without hindering the performance, flexibility, and applicability of the model-free RL algorithm.
We evaluate our approach on two representative dynamic robotic learning problems and find substantial performance improvement relative to previous curriculum generation techniques and na\"ive exploration strategies.

\end{abstract}

\section{Introduction}
\label{sec:intro}

Reinforcement learning (RL) is a powerful tool for training agents to maximize reward accumulation in sequential decision-making problems \cite{silver2016mastering,mnih2013playing,levine2016end, tan2018sim}. 
In particular, model-free approaches to robotic RL may allow for the completion of tasks that are difficult for traditional control theoretic tools to solve, such as learning policies mapping observations directly to actions \cite{levine2016end}. 
In spite of previous successes, RL has not seen widespread adoption in real-world settings, especially in the context of robotics. 
One of the fundamental barriers to applying RL in robotic systems is the extremely high sample complexity associated with training policies.


One approach to overcoming sample complexity is to train in simulation and port to the true environment, possibly with a ``sim-to-real'' algorithm to mitigate the impacts of simulator-reality mismatch \cite{tan2018sim, harrison2017adapt, tobin2017domain}.
While this is a promising approach that reduces the number of trials that are required to run on a physical robot, it still requires possibly millions of trials on simulated systems \cite{schulman2017proximal}. 
A root cause of this extremely high sample complexity is the use of na\"ive exploration strategies such as $\epsilon$-greedy  \cite{mnih2015human}, Ornstein-Uhlenbeck noise \cite{lillicrap2015continuous}, or in the case of stochastic policy gradient algorithms such as Proximal Policy Optimization (PPO) \cite{schulman2017proximal}, reliance on the stochasticity of the policy.
These dithering exploration strategies are well-known to take exponential time for problems with sparse reward \cite{osband2016deep}. 
Unfortunately, in robotic learning, smooth reward functions can lead to undesirable behavior and sparse rewards corresponding directly to accomplishing a task are often necessary \cite{thomas2018learning, florensa2017reverse}.
Moreover, if another component of the reward function is a cost associated with control, na\"ive exploration strategies may actually guide the system away from the goal region in an effort to minimize control costs. 

In the context of robotics, exploration efficiency has the potential to be substantially improved by leveraging physical priors \cite{bansal2017mbmf}. Model-based approaches to robotic RL have demonstrated exceptional sample efficiency \cite{deisenroth2011pilco}, but are often limited to generating local models and policies, and are not capable of solving as wide a variety of problems as model-free methods. Techniques to leverage models to guide exploration often require either substantial modification of the underlying model-free learning algorithm \cite{bansal2017mbmf, chebotar2017combining}, or leverage control theoretic tools which potentially prevents application to many interesting problems \cite{levine2014learning,gu2016continuous}.

\emph{Contributions:} We propose \algNameLong{} (\algName{}), a method of improving exploration efficiency in model-free RL by seamlessly leveraging approximate physical models without altering the RL algorithm.
Our approach works by altering the initial state distribution used when training an arbitrary model-free RL agent. We start training from states near the sparse goal region, and iteratively expand the initial state distribution by computing approximate backward reachable sets (BRSs): the set of all points in the state space capable of reaching a certain region in a fixed, short amount of time. 


By explicitly moving the initial state distribution backwards in time from the goal according to approximate dynamics of the system, we initially train on simple problems, ensuring the learning agent receives a strong reward signal. As the agent demonstrates mastery, we increase problem difficulty until arriving at the original learning problem. 
Our approach requires only approximate BRSs, so a low-fidelity approximate model can be used to compute them.
During forward learning, the original high-fidelity model is used, so our method is simply a wrapper that augments \emph{any} model-free RL algorithm. 

One key feature of our approach is that BRSs expand in all dynamically feasible directions of the state space, providing a ``frontier'' from which states are sampled.
By using prior knowledge of the system through approximate dynamics and performing backward reachability, we drastically improve sample efficiency while still learning a model-free control policy, thereby maintaining the potential to learn high quality policies with arbitrary inputs (such as images or other sensors inputs).
Additionally, our approach is capable of handling highly dynamic systems in which the forward and backward time dynamics vastly differ, and is modular in the sense that any model-free RL method and any reachability method can be combined.

We demonstrate our approach on a car environment and a dynamic planar quadrotor environment, and observe substantial improvements in learning efficiency over both standard model-free RL as well as existing curriculum approaches.
We further investigate a variety of scenarios in which there exists mismatch between the model used for curriculum generation and for forward learning, and find good performance even with model mismatch. 

\section{Related Work}
\label{sec:related}
One growing body of work that aims to improve exploration efficiency in general RL problems has been on the topic of \textit{deep exploration} \cite{osband2016deep}. 
Dithering strategies such as $\epsilon$-greedy aim to explore the state space by taking suboptimal actions with some low probability. 
However, this does not account for the relative uncertainty of value estimates, and thus an agent in an MDP with sparse reward will continuously explore a region around the initial state, as opposed to deliberately searching the state space \cite{osband2017deep}. 
General purpose deep exploration strategies address this problem by, for example, maintaining approximate distributions over value functions and sampling at the start of each episode \cite{osband2016deep}, or augmenting the MDP reward towards reducing the agent's uncertainty in environment dynamics \cite{houthooft2016vime}. 
In the specific context of robotic RL in simulation, the availability of physical models and the ability to adjust the parameters of the RL task while training should enable guiding exploration in a more direct manner; this is the high-level motivation for our work.

Our approach is based on ideas from curriculum learning, which aims to improve the rate of learning by first training on easier examples, and increasing the difficulty as training progresses until the original task becomes appropriate to train on directly \cite{bengio2009curriculum}. 
In supervised learning, a variety of hand-designed curricula methods as well as automatic curriculum generation methods have been effective in certain scenarios \cite{graves2017automated,zaremba2014learning}.
Several authors have also applied curriculum schemes to robotic systems \cite{karpathy2012curriculum,baranes2013active}; however, these approaches are based on learning pre-specified control primitives or identifying inverse dynamics models, as opposed to directly solving a given MDP. A common curriculum method in RL is reward ``smoothing'', in which a sparse reward signal is replaced with a smooth function, thus providing reward signal everywhere in the state space. A common example of this is providing a cost that is quadratic in the distance to the goal state. This approach has been shown to yield highly sub-optimal policies in numerous robotic control tasks \cite{thomas2018learning, florensa2017reverse}. For example, in a peg insertion task, \cite{thomas2018learning} found that the robot would instead place the peg beside the hole, as opposed to successfully inserting it. We compare against reward smoothing in our numerical experiments.

The work of \cite{florensa2017reverse} offers an attractive, purely model-free approach to curriculum generation for robotic RL, which adjusts problem complexity by adjusting the initial state distribution of the RL task during training. 
The authors argue that initial states which yield a medium success rate lead to good learning performance, and thus select new start states by sampling random actions from these medium success rate states.
This approach was applied successfully in a variety of systems, but can break down for highly dynamic or unstable systems. 
In such systems, the state reached by taking random actions from a ``good start state'' is likely to be far from the original state and thus unlikely to itself also be a ``good start state.'' 
Further, for systems where dynamics moving forward in time are very different from those moving backward in time, e.g. underactuated systems with drift, this random action sampling curriculum may never train from starts that belong to the true initial state distribution.
Compared to the work in \cite{florensa2017reverse}, our method can be viewed as a generalization that explicitly utilizes prior knowledge of a system. 
By expanding BRSs outward from the goal and generating a more uniform coverage of the state space, our approach effectively prioritizes exploration of the frontier of the BRS.

Backward reachability has been extensively used for verifying performance and safety of systems \cite{Bansal2017a, Chen2018}.
There are a plethora of tools for computing BRSs for many different classes of system models \cite{Mitchell07c,Althoff2015,Fan2016}.
Since the traditional focus of backward reachability has been on providing performance and safety guarantees, computations of BRSs are expensive.
Recent system decomposition techniques are effective in alleviating the computational burden in a variety of problem setups \cite{Mitchell03, Chen2016b, Chen2018b}.
As we will demonstrate, approximate BRSs computed using these decomposition techniques are sufficient for the purpose of guiding policy learning and do not significantly impact training time.



\section{Problem Formulation}

The goal of this work is to seamlessly leverage physical priors to provide a method for generating curricula to accelerate model-free RL for sparse-reward robotic tasks in simulation. In the following subsections we formalize the notion of a sparse-reward robotic RL task and specify the physical priors we leverage.







\subsection{Sparse Reward MDP}
We assume the robotic RL task is defined as a discrete time MDP of the form $(\mathcal{S},\mathcal{A},F,r,\rho_0,\gamma,\mathcal{S}_g,\mathcal{S}_f)$, in which $\mathcal{S} \subset \mathbb{R}^{n}$ and $\mathcal{A} \subset \mathbb{R}^{m}$ are the state and action sets respectively, $\rho_0$ is the distribution of initial states, and $F: \mathcal{S} \times \mathcal{A} \to \mathcal{S}$ is the transition function which may be stochastic.

The MDP is an infinite horizon process with discount factor $\gamma$. There exist two sets of absorbing states which we refer to as \textit{goal states}, $\mathcal{S}_g \subset \mathcal{S}$, and \textit{failure states}, $\mathcal{S}_f \subset \mathcal{S}$. 
This distinction is for technical reasons, but is an intuitive one that is common in robotic motion planning \cite{lavalle2006planning}. 
For example, a goal state may be a robotic arm successfully placing an object in the correct location, whereas a failure state may be the robot dropping the object out of reach. 

The function $r: \mathcal{S} \times \mathcal{A} \to \mathbb{R}$ is a reward function, composed of a strictly negative running cost\footnote{We use the term running cost to denote a cost associated with a non-terminal state-action pair.}, and the cost or reward associated with the goal and failure regions. 
The running cost implies that a robot can only accumulate negative reward during normal operation, and thus makes it desirable to reach the goal region as opposed to continuously accumulating reward.
Finally, we make the technical assumption that for the optimal policy, for all $s \in \mathcal{S}$, trajectories terminating in the goal region have higher expected reward than those terminating in the failure region. 
This is simply saying that it is always desirable to try to reach the goal region and to avoid failure. 
The reward structure presented herein is a generalization of that presented in \cite{florensa2017reverse}, and allows us to incorporate cost components that are common and desirable in robotics such as control effort penalties.

We assume we are training in simulation, and thus we can set the initial state of each episode arbitrarily. Furthermore, we assume we have knowledge of the goal region.


\subsection{Approximate Dynamics Model as a Physical Prior} \label{sec:cdyn}


We assume access to physical prior knowledge in the form of a simplified, approximate dynamics model of the system, henceforth referred to as the \textit{curriculum model} $\hat{M}$, with which approximate BRSs can be efficiently computed to guide the policy learning process. 
Depending on the method used to compute BRSs, an appropriate choice of curriculum model could take the form of an ODE, a difference equation, or an MDP.
In this paper, we choose to use the Hamilton-Jacobi (HJ) reachability formulation, and hence assume that this curriculum model takes the form of a continuous-time ODE. 
In accordance with HJ reachability, let the curriculum state be denoted $\cstate\in \hat{\mathcal{S}} \subset \mathbb{R}^{\hat n}$, with dynamics given by
\begin{align}
\label{eq:cdyn}
\dot\cstate(t) = \cdyn(\cstate(t), \cctrl(t)), \,\, \cctrl(t) \in \ccset,
\end{align}
where the simulator state dimension $n$ is not necessarily equal to $\hat{n}$.
Specifically, we define a injective map $\scmap(\cdot): \R^n \rightarrow \R^{\hat n}$ that maps the simulator model state to the curriculum model state.
In general, $\scmap(\cdot)$ is nonlinear and often a projection. The curriculum dynamics should be chosen to be a minimally sufficient representation of the dynamics, and is only need for a coarse approximation of the BRSs.

The curriculum dynamics $\cdyn: \hat{\mathcal{S}} \times \ccset \rightarrow \hat{\mathcal{S}}$ are assumed to be uniformly continuous, bounded, and Lipschitz continuous in $\cstate$ for fixed $\cctrl$, so that given a measureable control function $\cctrl(\cdot)$, there exists a unique trajectory solving \eqref{eq:cdyn} \cite{Coddingtona}. 







\section{Approach}
\label{sec:approach}
\newcommand{\BRS}{\mathrm{BRS}_T}
Our full approach is summarized in Algorithm \ref{alg:full_alg}\footnote{The code used in this paper is available at \url{https://github.com/StanfordASL/BaRC}}. The intuition behind our method stems from pedagogy, where learners are taught foundational topics that are later layered with advanced study. 
BRSs offer a convenient choice as foundational topics in the context of continuous control tasks---in order to reach a state $s$ in time $T$, a trajectory must pass through the BRS with time horizon $T$ of state $s$, which we denote as $\maxbrs(T; \state)$.  
Thus, our algorithm begins by requiring the learner (an RL policy) to learn how to transition from states in the goal's BRS to the goal. Once the learner can successfully reach the goal from these states,
the curriculum computes their BRS
and uses this expanded set as new initial states from which to train, thus incrementally increasing the difficulty of the task.
This process of expanding the initial-state distribution in a dynamically-informed manner continues iteratively until the BRS spans the state space or a given start state is reached. 
The rest of this section describes our algorithm in detail.

Every stage of the curriculum is defined by determining the set of initial states to train from by calling \textsc{ExpandBackwards} to compute starts\_set, the union of $\maxbrs(T; \cstate)$ for every $\cstate$ in starts, which initially just contains a state in the goal region.
The inner loop of the algorithm represents training a policy on this stage of the curriculum.
Rather than only training on states from this expanded set, we sample $N_\text{new}$ states from the start\_set and mix in $N_\text{old}$ states sampled from old\_starts, a list of states from which the policy has previously demonstrated mastery of the task. In this way, we avoid catastrophic forgetting during the policy training. 
We train the policy using a model-free policy optimization algorithm for $N_\text{TP}$ iterations via the \textsc{TrainPolicy} subroutine. The subroutine also returns the success rate of the policy from the different initial states sampled during training.
The starts list is updated to contain initial states from which the policy can reliably reach the goal. This is done via the \textsc{Select} subroutine, which returns initial states in success\_map with success rate greater than $C_\text{select}$. These states are also added to the old\_starts buffer.


This inner loop repeats until the call to \textsc{Evaluate} determines that the fraction of states in this stage of the curriculum from which $\pi$ can reach the goal exceeds $C_\text{pass}$. 
At this point, $\pi$ is considered to have adequately mastered this curriculum stage, and we move to the next curriculum stage.

\paragraph*{Hyperparameters} 
The algorithm has six hyperparameters. $C_\text{select}$ and $C_\text{pass}$ define a notion of mastery from a particular state and a curriculum stage respectively. The horizon $T$ used in the BRS computation controls how much the task difficulty increases between stages of the curriculum. The ratio of $N_\text{old}$ to $N_\text{new}$ balances training on new scenarios with preventing forgetting. The number of training iterations between evaluations $N_\text{TP}$ should ideally be set to a value such that the policy has a high chance of mastering a stage of the curriculum after $N_\text{TP}$ iterations of the policy optimization algorithm, in order to avoid unnecessary evaluation checks. Empirically, the algorithm is robust to the settings of these hyperparameters.

\begin{algorithm}[t]
\begin{algorithmic}
\small
\Require Hyperparameters $N_\text{new}, N_\text{old}, T, C_\text{pass}, N_\text{TP}, C_\text{select}$
\Function{BaRC}{$s_g$, $\rho_0$, $\hat{M}$}
\State $\pi \gets \pi_0$
\State starts $\gets$ $\{s_g\}$
\State oldstarts $\gets \{s_g\}$
\For{$i$ = 1, 2 \dots}
\State starts\_set $\gets$ \Call{ExpandBackwards}{starts, $\hat{M}$, $T$}
\State frac\_successful $\gets$ 0.0
\While{frac\_successful $< C_\text{pass}$}
\State $\rho_i \gets \mathrm{Unif}(\text{starts\_set}, N_\text{new}) \cup \mathrm{Unif}(\text{oldstarts}, N_\text{old})$ 
\State $\pi$, success\_map $\gets$ \Call{TrainPolicy}{$\rho_i$, $\pi$, $N_\text{TP}$}
\State starts $\gets$ \Call{Select}{success\_map, $C_\text{select}$}
\State oldstarts $\gets$ oldstarts $\cup$ starts
\State frac\_successful $\gets$ \Call{Evaluate}{$\rho_i$, $\pi$}
\EndWhile
\State iter\_result $\gets$ \Call{PerformanceMetric}{$\rho_0$, $\pi$}
\EndFor
\State \Return $\pi$
\EndFunction
\end{algorithmic}
\caption{The full BaRC algorithm.}
\label{alg:full_alg}
\end{algorithm}


\paragraph*{Computing BRSs}
To actually obtain BRSs, we use methods from reachability analysis. 
Specifically, we use the HJ formulation of reachability since our approach focuses on capturing key nonlinear behaviors of systems (through the curriculum dynamics model). 
The BRS of a target set $\targetset \subset \R^{\hat n}$ represents the set of states of the curriculum model $\cstate\in\R^{\hat n}$ from which the curriculum system can be driven into $\targetset$ at the end of a short time horizon of duration $T$. In our work, the BRS of a target set $\targetset \subset \R^{\hat n}$, which we denote $\maxbrs(T; \targetset)$, is formally defined to be
\begin{equation*}
 \left\{\cstate_0: \exists \cctrl(\cdot), \cstate(\cdot) \text{ satisfies \eqref{eq:cdyn} },\cstate(-T) = \cstate_0, \cstate(0) \in \targetset \right\},
\end{equation*}
which is obtained as the zero sublevel set of a value function $\valfunc(-T, \cstate)$ that is the solution to an HJ partial differential equation (PDE)~\cite{Bansal2017a, Chen2018}: $\maxbrs(T; \targetset) = \{\cstate: \valfunc(-T, \cstate) \le 0\}$.
More details are given in the appendix.
When the target set consists of a single state $\cstate$, we denote the BRS $\maxbrs(T; \cstate)$.

Solving the PDE is computationally expensive in general.
However, since here we are only using BRSs to guide policy search rather than their traditional application of verifying system performance and safety, we can make approximations. 
Specifically, we utilize system decomposition methods \cite{Mitchell03, Chen2016b, Chen2018b} with the simplified curriculum model in \eqref{eq:cdyn} to obtain approximate BRSs without significantly impacting the overall policy training time. 
The techniques in \cite{Mitchell03, Chen2016b, Chen2018b} provide outer approximations, although an outer approximation is not necessary for guiding policy search; any approximation that captures key nonlinear system behavior suffices.
Each system we experiment on is first decomposed into overlapping subsets of one or two components, which can each be solved efficiently and run online in our algorithm. Specifically, we use the open source helperOC\footnote{Found at \url{https://github.com/HJReachability/helperOC}} and Level Set Methods\footnote{Found at \url{http://www.cs.ubc.ca/~mitchell/ToolboxLS/}} toolboxes in MATLAB.

\paragraph*{Sampling from a BRS}
We perform rejection sampling over the components of the decomposed BRS approximation, meaning each component has low dimensionality (typically 1-D or 2-D) and thus it is inexpensive to evaluate membership. We first determine a tight bounding box around each component's BRS (this can be computed efficiently from widely available low-level contour plotting methods, e.g. \texttt{contourc} in MATLAB). Then, we uniformly sample points in the bounding box and reject any that fail membership checks. The initial bounding box calculation enables us to efficiently perform rejection sampling in cases where the size of the BRS is much smaller than the state space. Visualizations of our rejection sampling method and of BRSs can be found in the appendix.

While a more formal theoretical analysis is out of the scope of this paper, the theoretical benefits of sampling based on BRSs can be gleaned from the definition of the BRS, which involves an existential quantifier for the control function $\cctrl(\cdot)$.
This implies that a state that can possibly reach the goal using \textit{any} policy is included in the BRS.
This is in contrast to methods that sample the state space through applying random controls.



\begin{figure}[t]
\centering
\includegraphics[width=0.9\linewidth]{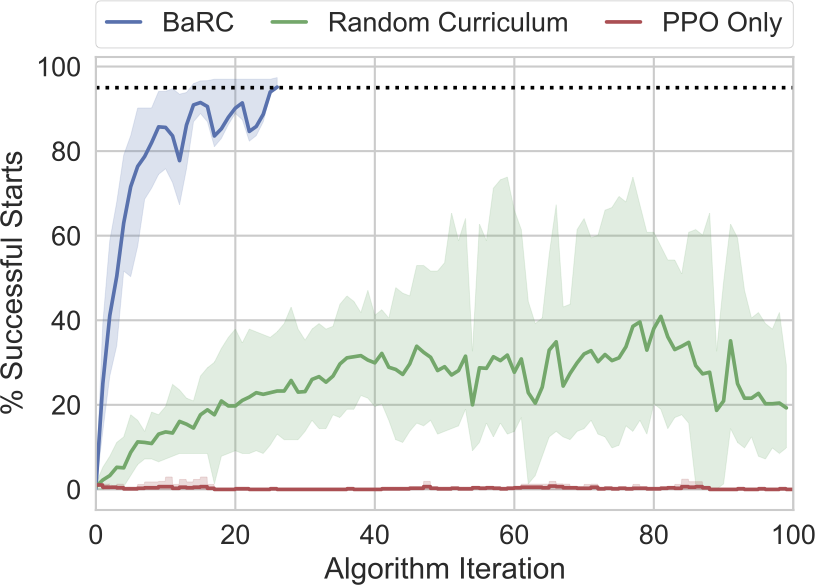}
\includegraphics[width=0.9\linewidth]{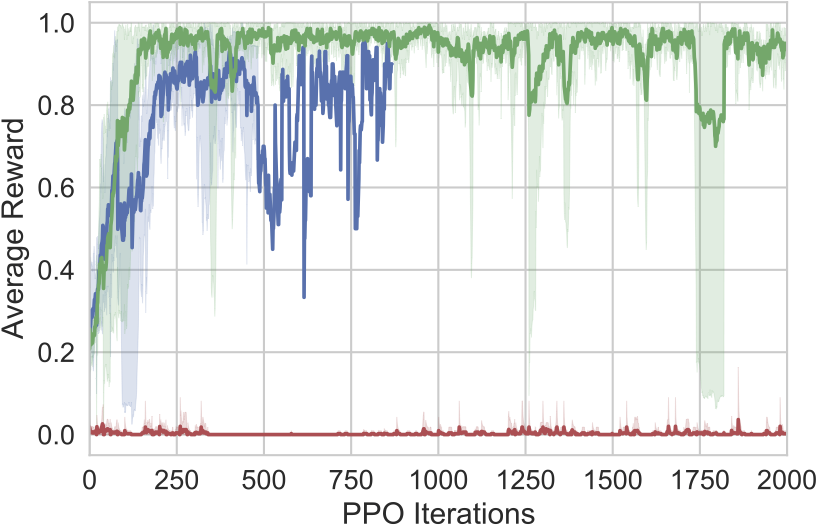}
\caption{Performance of the car model under \algName{}, the approach of \cite{florensa2017reverse} (referred to as Random Curriculum), and standard PPO \cite{schulman2017proximal}. \textbf{Upper:} Percentage of start states in which the learned policy can reach a specified goal state. This curve is equivalent to expected reward over a wide initial state distribution. Mean and 95\% confidence intervals were computed from the results of 5 different runs. \textbf{Lower:} Average reward during policy training. Note that it is \textit{not} desirable to have expected reward be approximately 1, as this implies that the curriculum is not providing sufficiently challenging problems. Instead, it is desirable to ensure the reward is between 0 and 1 to provide a learning signal for the agent during training.}
\label{fig:no_err_expts}
\end{figure}

\begin{figure*}[t]
\centering
\includegraphics[width=0.325\linewidth]{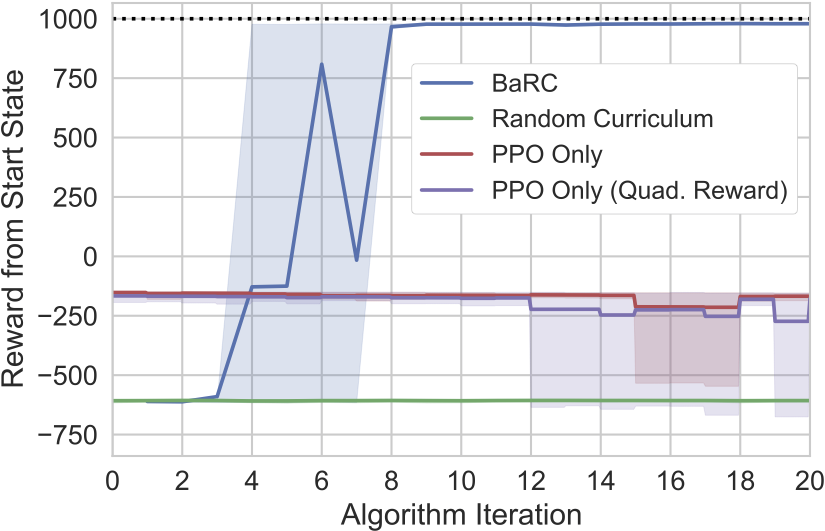}
\includegraphics[width=0.325\linewidth]{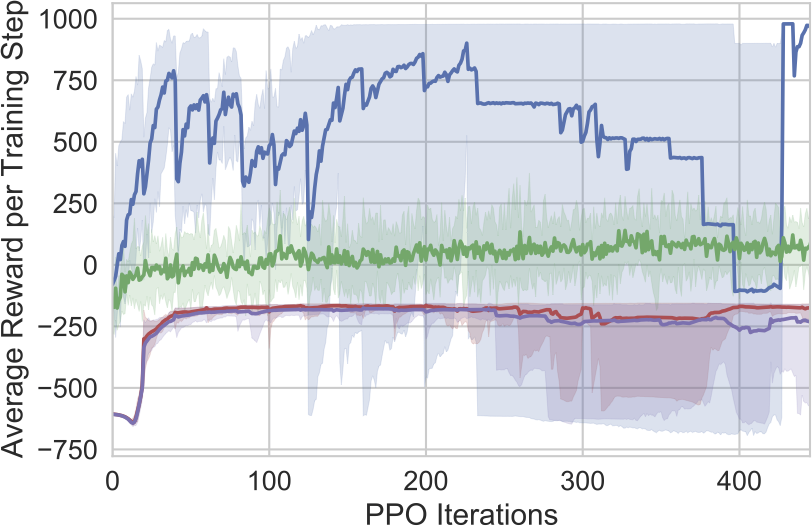}
\includegraphics[width=0.325\linewidth]{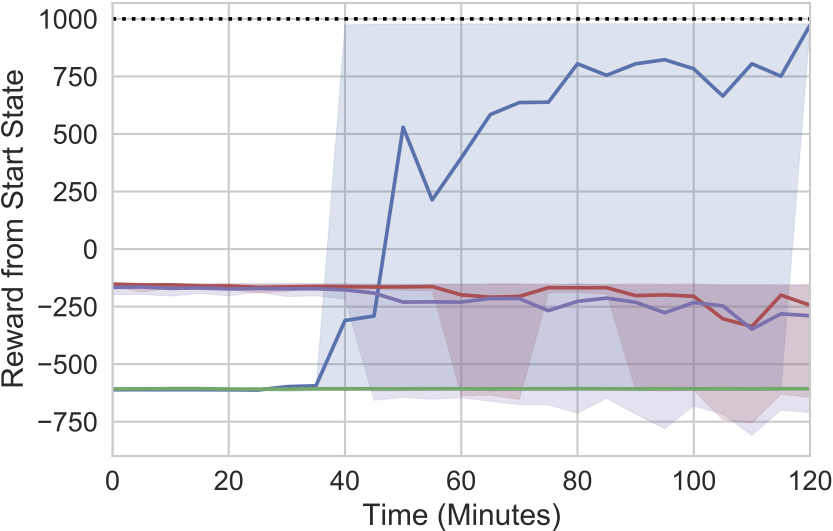}
\caption{Performance of \algName{}, random curriculum, and standard PPO on the planar quadrotor task. \textbf{Left:} Mean reward over ten experiments for the three approaches (95\% confidence intervals). Note that between 4-8 iterations, the different instantiations of \algName{} learn to reach the goal. The PPO policy simply learns how to however, whereas the random curriculum never learns any non-colliding behavior. \textbf{Center:} The average reward during training of the policy. The iterative increasing and decreasing of average reward for the \algName{} agent shows the policy learning for a given BRS, followed by \algName{} expanding the BRS. This alternating learning and expansion is repeated until the \algName{} agent reaches the initial state and the algorithm achieves maximal reward in every iteration. \textbf{Right:} The mean reward over ten experiments, plotted versus wall clock time. }
\label{fig:quad_expts}
\end{figure*}

\section{Experimental Results}
\label{sec:result}

We perform numerical experiments on two dynamical systems, and evaluate a variety of model mismatch scenarios to investigate the robustness of \algName{} to inaccurate curriculum models. All experiments were performed using the PPO algorithm as the model-free policy optimization method. The hyperparameters for both experiments are listed in the appendix.

\subsection{\expA}

We use a five dimensional car that is a standard test environment in motion planning \cite{webb2013kinodynamic} and RL \cite{harrison2017adapt}. The car has state $s = [x,y,\theta,v,\kappa]^T$, where $x$ and $y$ denote position in the plane, $\theta$ denotes heading angle, $v$ denotes speed, and $\kappa$ denotes trajectory curvature. The system takes as input $a = [a_v,a_\kappa]^T$, where $a_v$ is longitudinal acceleration and $a_\kappa$ is the derivative of the curvature. The reward function is an indicator variable for a small region around the goal state. The goal state is a point at the origin with velocity sampled uniformly from 0.1 to 1.0 m/s and heading sampled uniformly from $-\pi$ to $\pi$ radians. We consider states that are within 0.1 m in distance, 0.1 m/s in velocity, and 0.1 radians of the goal state to have reached the goal. The agent receives a reward of 1.0 upon reaching the goal, and 0.0 otherwise. The simulator model was used as a curriculum model for the experiments presented in the body of the paper. 
The appendix
contains a collection of results for cases where there exists a mismatch between the simulator model and the curriculum model, and these systems performed similarly to the cases presented here.

To test coverage of the state space via the \algName{} algorithm, we initialized a set of 100 points ``behind" the goal state. Here, ``behind" means that any sampled point's projection onto the goal state's velocity vector is negative. Further, the maximum angle allowed between any sampled state and the goal state's negative velocity vector is $\pi/4$. Finally, we orient the sampled states randomly within $\pi/4$ radians of the goal vector's orientation. Each of the sampled states has zero initial velocity and curvature. We then ran \algName{}, evaluating at each iteration the number of original sampled states from which the learned policy could traverse to the goal. We chose this set as it is representative of states from which a policy could realistically traverse to a goal with nonzero velocity.

Figure \ref{fig:no_err_expts} shows the performance of \algName{}, as well as the curriculum method presented in \cite{florensa2017reverse} (which we refer to as \textit{Random Curriculum}) and standard PPO \cite{schulman2017proximal}. 
The top plot shows the number of start states from which the algorithm successfully reaches the goal state. 
When the success rate exceeded 95\%, the task was considered solved and training was terminated. \algName{} achieves rapid coverage of the state space and a success rate of greater than 95\% for all five experiments in under thirty iterations of the outer loop. 
In contrast, PPO obtains almost no reward. 
The random curriculum approach successfully connects for between twenty and forty percent of start states and shows little improvement beyond the first 40 iterations. 
On the bottom, the average reward during training is plotted. 
These reward returns are for training from curriculum-selected start states as opposed to the true initial state distribution of the problem. 
As a result, counter-intuitively, it is desirable to have neither an average reward close to zero (implying an ineffective exploration scheme), nor an average reward close to one (implying the curriculum is expanding too slowly). 
The average reward for \algName{} varied between approximately 0.4 and 0.9, demonstrating that the algorithm creates a well-paced curriculum that ensures good, continuous learning progress.

\subsection{\expB}



\begin{figure*}[t]
\centering
\includegraphics[width=0.245\linewidth]{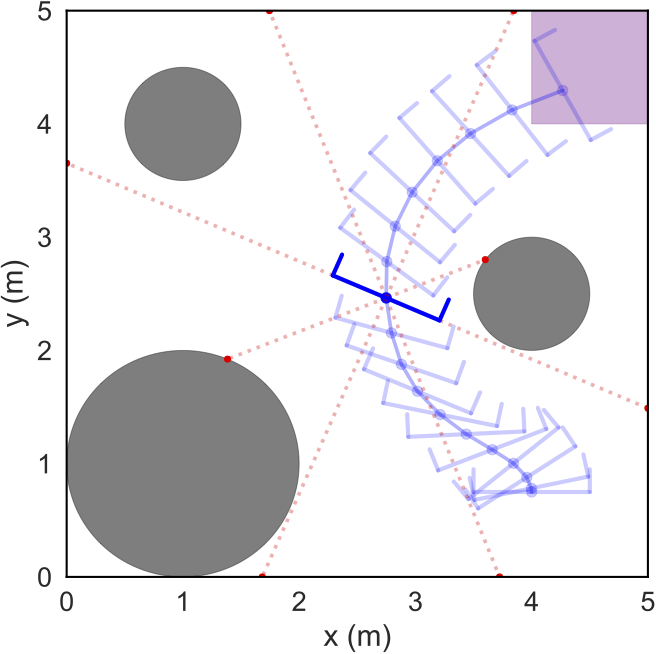}
\includegraphics[width=0.245\linewidth]{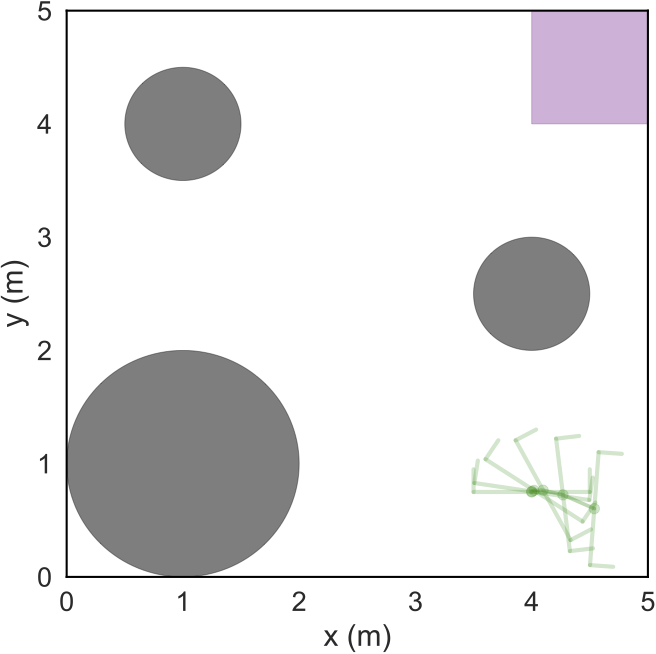}
\includegraphics[width=0.245\linewidth]{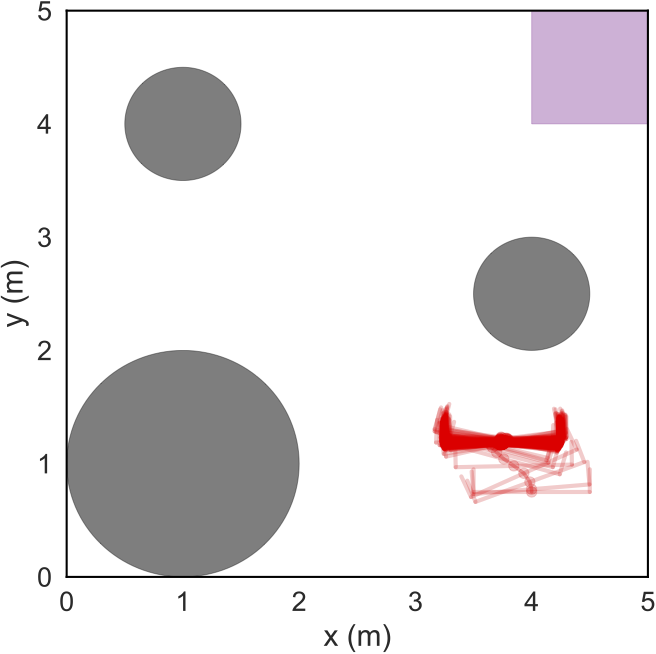}
\includegraphics[width=0.245\linewidth]{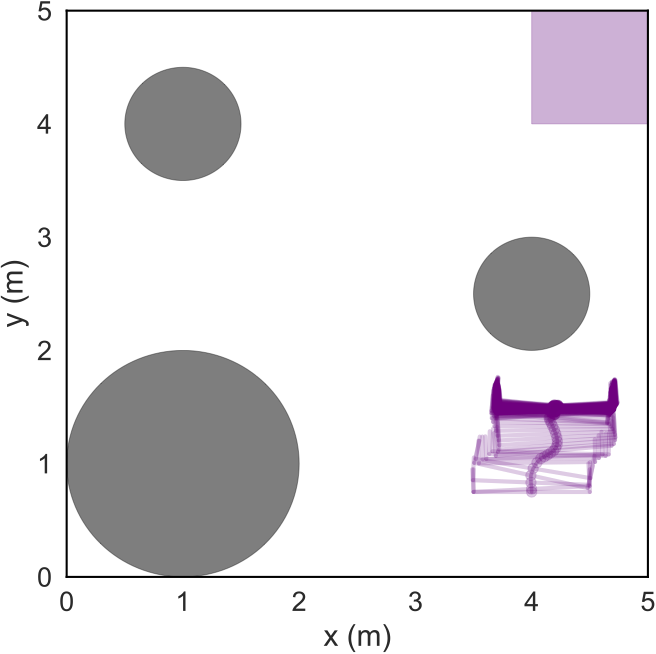}
\caption{\textbf{Left:} Visualization of a trajectory from a BaRC-trained policy. The policy was trained for 10 BaRC iterations. The red lines denote laser rangefinder measurements which were included in the observation. They are excluded from the other two plots for clarity. Magenta indicates the goal region and grey indicates obstacles. \textbf{Center Left:} A trajectory from a policy trained using the random curriculum approach from \cite{florensa2017reverse}. It was trained for 10 curriculum iterations. \textbf{Center Right:} A trajectory from a policy trained using standard PPO \cite{schulman2017proximal}. It was trained for 200 PPO iterations (equivalent to 10 outer loops of 20 PPO iterations). \textbf{Right:} A trajectory from a policy trained using standard PPO, with a quadratized reward function.}
\label{fig:quad_traj_comparison}
\end{figure*}

To evaluate the performance of \algName{} on a highly dynamic, unstable system, we tested the algorithm on a planar simulation of a quadrotor flying in clutter. This planar quadrotor is a standard test problem in the control literature \cite{gillula2010design,singh2017robust}. This system has state $s = [x, v_x, y, v_y, \phi, \omega]$, where $x,y,\phi$ denote the planar coordinates and roll, and $v_x, v_y, \omega$ denote their time derivatives. The actions available to the agent are the two rotor thrusts, $T_1$ and $T_2$. The learning agent receives an observation that is augmented with 8 laser rangefinder sensors (see Figure \ref{fig:quad_traj_comparison} for a visualization). These sensors are placed at every 45 degrees and provide the distance from the vehicle to the nearest obstacle. This observation allows the learning agent to map directly from sensor inputs to actions.

In this problem, the agent receives a reward of 1000 for reaching the goal region (defined as $\mathcal{S}_g = \{s : x\geq4, y\geq 4\}$) and accumulates a control cost of $0.01(T_1^2 + T_2^2)$ per timestep. The episode runs for at most 200 timesteps, and the maximum thrust is set to be twice the weight of the quadrotor. If the agent collides with an obstacle or the walls it receives a cost equal to the maximum control cost over the entire episode. As such, it is always desirable for the agent to avoid collisions. Because of the extremely sparse reward associated with reaching the goal, the unstable dynamics, and absorbing collision states, this problem is extremely difficult to solve for standard exploration schemes.

Figure \ref{fig:quad_expts} shows the performance of \algName{} as well as the random curriculum and standard PPO.
PPO, as expected, never reaches the goal and thus cannot learn to deliberately move toward the goal region. 
Instead, it learns a locally optimal policy which hovers in place to avoid collision. We also compare to standard PPO with a smoothed reward function. In this version of the problem, the reward function is augmented with a quadratic cost term, penalizing distance from the goal. As a result, the learned policy is a highly sub-optimal local minimum, in which the agent hovers slightly closer to the obstacle above it, and to the right. It does not, however, learn a policy that allows it to reach the goal.

The random curriculum, because it takes random actions forward and backward, rapidly becomes unstable and achieves only infrequent rewards. 
It does not advance far enough in the state space to produce any deliberate goal-seeking behavior from the true start state of the problem. 
In contrast, the \algName{} agent learns to move to the goal every time within 4-8 iterations of the curriculum. 
As can be seen in the average reward plot (Figure \ref{fig:quad_expts}, center), there are discrete waves of policy improvement (average reward increase) followed by iterative expansion of the BRS (visible by the associated sharp reward decrease). 
This continues until the curriculum reaches the initial state. It then rapidly trains from the initial state, quickly improving to be able to consistently solve the problem. 
Sample trajectories are visualized in Figure~\ref{fig:quad_traj_comparison}. 
Because we do not enforce a hover condition in the goal region, the \algName{} policy learns extremely aggressive trajectories toward the goal.

Since PPO exploration steps are interleaved with expansion of the BRS, the average time per iteration is slower than standard PPO or the random curriculum. The performance versus wall clock time for both the car system and the planar quadrotor is presented in Figure \ref{fig:quad_expts} (right). Each iteration of \algName{} takes roughly 2-3 times as long as standard PPO iterations. Even considering performance versus clock time, \algName{} demonstrates dramatic performance improvement versus previous approaches. Moreover, once the BRS expands to include the initial state, expansion of the BRS may be halted and training can continue without any curriculum computation overhead. 


\section{Discussion and Conclusions}
\label{sec:conclusion}



In this paper, we proposed backward reachability curriculum (\algName{}), which addresses the challenge of sample complexity in model-free reinforcement learning (RL) by using backward reachable sets (BRSs) to provide curricula for the policy learning process.
In our approach, we initially train the policy starting close to the goal region, before iteratively computing BRSs using a simplified dynamic model to train the policy from more difficult initial states.
\algName{} is an effective way to leverage physical priors to dramatically increase the rate of training of model-free algorithms without affecting their flexibility in terms of applicability to a large variety of RL problems, and enables the solution of sparse reward robotic problems that were previously intractable with standard RL exploration schemes. The hyperparameters of \algName{}, such as the time horizon of BRS expansions and the mastery threshold, are intuitive to a system designer and may be easily adjusted. 
In our numerical examples, since we used continuous time BRS methods, the time horizon of BRS expansion may be chosen to be any positive number, as opposed to being limited to the same time discretization of the simulator.
Discrete time BRS methods can also be used to a similar effect, since the time discretization of the curriculum model need not be the same as that of the simulator model.

While we have seen strong exploration efficiency gains from directly computing BRSs, it is unclear if our approach could be relaxed to a sampling based scheme. Our approach mimics dynamic programming, first solving the control subproblem close to the goal region, and iteratively increasing the problem complexity while relying on the solutions to the previously solved subproblems. Sampling based analogues to dynamic programming exist in, for example, motion planning \cite{janson2015fast, ichter2017group}. These sampling based algorithms, using the curriculum dynamics, may be a promising way to increase the efficiency of approximating BRSs. 

There are several additional avenues of future work. First, incorporating both forward and backward reachability to speed up policy training when a particular initial state distribution is of interest has to potential to improve training efficiency. Investigating potential performance benefits of tuning the BRS time horizon and quantifying the sample complexity benefits of using BRSs versus simulating trajectories potentially with respect to the quality of the curriculum model are important characterizations of \algName{}.
Finally, performing efficient system identification \cite{ljung1998system,levine2013guided,harrison2018meta} during portions of the training process that occur on the true to obtain a curriculum model in situations when an approximate model is not known a priori may be investigated to increase the set of problems \algName{} may be applied to.







\section*{Acknowledgment}

The authors were partially supported by the Office of Naval Research, ONR YIP Program, under Contract N00014-17-1-2433 and the Toyota Research Institute (``TRI"). James Harrison was supported in part by the Stanford Graduate Fellowship and the National Sciences and Engineering Research Council (NSERC). This article solely reflects the opinions and conclusions of its authors and not ONR, TRI or any other Toyota entity.


\bibliographystyle{IEEEtran}

\bibliography{ref}  


\appendix
\section{Backward Reachability}
\subsection{Computing backward reachable sets (BRSs)}
Intuitively, a BRS represents the set of states of the curriculum  model $\cstate\in\R^{\hat n}$ from which the curriculum system can be driven into a target set $\targetset \subset \R^{\hat n}$ at the end of a time horizon of duration $T$. 

In this paper, we focus on the ``Maximal BRS\footnote{The term ``maximal'' refers to the role of the optimal control, which is trying to maximize the size of the BRS.}
''; in this case the system seeks to enter $\targetset$ using some control function, and the Maximal BRS (referred to as simply ``BRS'' in this paper) represents the set of states from which the system is guaranteed to reach $\targetset$.

\begin{defn}
  \label{defn:rset_goal}
  \textbf{Backward Reachable Set (BRS)}.
  \begin{equation*}
  \maxbrs(T; \targetset) = \{\cstate_0: \exists \cctrl(\cdot), \cstate(\cdot) \text{ satisfies \eqref{eq:cdyn}},\cstate(-T) = \cstate_0, \cstate(0) \in \targetset \}
  \end{equation*}
\end{defn}

HJ formulations outlined in the review papers \cite{Bansal2017a, Chen2018} cast the reachability problem as an optimal control problem and directly compute BRSs in the full state space of the system. 
The numerical methods such as those implemented in \cite{Mitchell07c} for obtaining the optimal solution all involve solving an HJ PDE on a grid that represents a discretization of the state space.

Let the target set $\targetset \subset \R^{\hat n}$ be represented by the implicit surface function $\fc(\state)$ as $\targetset = \{\cstate: \fc(\cstate) \le 0\}$. 
Consider the optimization problem

\begin{equation}
\label{eq:ocpplus}
\begin{aligned}
\valfunc_\maxbrs(t, \state) = \inf_{\cctrl(\cdot)} \fc(\cstate(0)) \quad \text{subject to \eqref{eq:cdyn}}
\end{aligned}
\end{equation}

The value function $\valfunc_\maxbrs(-T, \cstate)$ is given by the implicit surface function representing the maximal BRS: $\maxbrs(T; \targetset) = \{\state: \valfunc_\maxbrs(-T, \state) \le 0 \}$, and is the viscosity solution \cite{Crandall1983} of the following HJ PDE, derived through dynamic programming:

\begin{align}
\label{eq:set_hj}
\frac{\partial \valfunc(t, \cstate)}{\partial t}  + \ham(\cstate, \nabla\valfunc(t, \cstate)) &= 0, \quad t \in [-T, 0], \\
\valfunc(0, \cstate) &= \fc(\cstate),
\end{align}

\noindent where the Hamiltonian $\ham(\state, \costate)$ is given by

\begin{equation}
\label{eq:ham_plus}
\ham(\cstate, \costate) = \min_{\cctrl \in \ccset} \costate \dyn (\cstate, \cctrl).
\end{equation}

\subsection{System decomposition}
Solving the HJ PDE \eqref{eq:set_hj} is computationally expensive; BRSs for systems with more than 5 state dimensions are computationally intractable using the HJ formulation.
In order to take advantage of the HJ formulation's ability to compute BRSs for nonlinear systems without significantly impacting the overall policy training time, we will utilize system decomposition methods to obtain approximate BRSs. 
In particular, in \cite{Chen2018b}, the authors proposed decomposing systems into ``self-contained subsystems'', and in \cite{Mitchell03} and \cite{Chen2016b}, the authors proposed ways to compute projections of BRSs.
We will combine these methods to obtain over-approximations of BRSs for the curriculum model.
Specific details of each curriculum model will be described in Section \ref{sec:exp_details}.

It is important to note that in this paper BRSs are used to guide policy learning, and as we will demonstrate, approximate BRSs computed with the simplified curriculum model in \eqref{eq:cdyn} are sufficient for the purposes of this paper.

\section{Experimental Details} \label{sec:exp_details}
\subsection{\expA}
For the car environment, we used the following five-state simulator model:
\begin{equation}
\label{eq:5D_dyn}
\dot \state = 
\begin{bmatrix}
\dot x\\
\dot y\\
\dot\theta\\
\dot v\\
\dot \kappa
\end{bmatrix} =
\begin{bmatrix}
(v + d_v) \cos \theta \\
(v + d_v) \sin \theta\\
\omega \\
a_v + d_{v-\text{ctrl}}\\
d_{a_\kappa} (a_\kappa + d_{\kappa-\text{ctrl}})
\end{bmatrix},
\quad |a_v| \le \bar a_v, |a_\kappa|\le\bar a_\kappa,
\end{equation}

\noindent where the position $(x,y)$, heading $\theta$, speed $v$, and angular speed $\kappa$ are the car's states.
The control variables are the linear acceleration $a_v$ and angular acceleration $a_\kappa$.
The variables $d_v, d_{a_v}, d_{a_\kappa}$ represent noise in the speed, linear acceleration, and angular acceleration, and are stochastic.
We assume that the learning agent has full access to the car state.

For the curriculum model, we used the same state variable and dynamics, but without the disturbance variables:
\begin{equation}
\label{eq:5D_cdyn}
\dot \cstate = 
\begin{bmatrix}
\dot{\hat x}\\
\dot{\hat y}\\
\dot{\hat\theta}\\
\dot{\hat v}\\
\dot{\hat\kappa}
\end{bmatrix} =
\begin{bmatrix}
\hat v \cos \hat\theta \\
\hat v \sin \hat\theta\\
\hat\kappa \\
\hat a_v \\
\hat a_\kappa
\end{bmatrix},
\quad |\hat a_v| \le \hat{\bar{a}}_v, |\hat{a}_\kappa|\le\hat{\bar{a}}_\kappa.
\end{equation}

In order to ensure that the approximate BRS computation can be done quickly, we decomposed \eqref{eq:5D_cdyn} as follows:

\begin{subequations} 
\label{eq:5D_cdyn_decomp}
\begin{align}
\begin{bmatrix}
\dot{\hat x}\\
\dot{\hat\theta}
\end{bmatrix} &=
\begin{bmatrix}
\hat v \cos \hat\theta \label{eq:5D_cdyn_decomp_XT}\\
\hat\kappa
\end{bmatrix},
&\underline\kappa \le \hat\kappa \le \bar\kappa, \quad \underline v  \le \hat v\le\bar v \\
\begin{bmatrix}
\dot{\hat y}\\
\dot{\hat\theta}
\end{bmatrix} &=
\begin{bmatrix}
\hat v \sin \hat\theta \label{eq:5D_cdyn_decomp_YT}\\
\hat\kappa
\end{bmatrix},
&\underline\kappa\le \hat\kappa \le \bar\kappa, \quad \underline v  \le \hat v\le\bar v \\
\dot{\hat v} &= \hat{a}_v,
&|\hat{a}_v| \le \hat{\bar{a}}_v, \label{eq:5D_cdyn_decomp_V} \\
\dot{\hat\kappa} &= \hat{a}_\kappa, 
&|\hat{a}_\kappa|\le\hat{\bar{a}}_\kappa \label{eq:5D_cdyn_decomp_W}
\end{align}
\end{subequations}

The decomposition occurs in two steps. 
First, the 5D curriculum dynamics are decomposed into two 4D subsystems with states $(\hat x, \hat \theta, \hat v, \hat \kappa)$ and $(\hat y, \hat \theta, \hat v, \hat \kappa)$.
This results in an over-approximation of the BRS \cite{Chen2018b}.
Next, $\hat v$ and $\hat\kappa$ are separated from $(\hat x, \hat \theta)$ and $(\hat y, \hat \theta)$, resulting in the subsystems in \eqref{eq:5D_cdyn_decomp}. In \eqref{eq:5D_cdyn_decomp_XT} and \eqref{eq:5D_cdyn_decomp_YT}, $\hat v$ and $\hat\kappa$ become fictitious control variables that are dependent on the subsystem BRSs in \eqref{eq:5D_cdyn_decomp_V} and \eqref{eq:5D_cdyn_decomp_W}, which results in over-approximation of the BRS according to \cite{Mitchell03, Chen2016b}.

\paragraph*{Hyperparameter Settings} In all experiments involving the car environment, we chose
\begin{align*} 
&N_\text{new} = 200,\ N_\text{old} = 100,\ T = 0.1 s,\\ 
&C_\text{pass} = C_\text{select} = 0.5,\ N_{\text{TP}} = 20
\end{align*}
These values were chosen with almost no tuning effort, as \algName{} is robust to hyperparamter choice. 

\subsection{\expB}

For the planar quadrotor environment, we used the same dynamics for the simulator and curriculum models, given by 

\begin{subequations} 
\label{eq:quad6D}
\begin{align}
	\begin{bmatrix}
	\dot x\\
	\dot v_x\\
	\dot y\\
	\dot v_y\\
	\dot \phi\\
	\dot \omega
	\end{bmatrix}
	=
    \begin{bmatrix}
	v_x\\
	-\frac{1}{m}C^v_D v_x -\frac{T_1}{ m}\sin\phi -\frac{T_2}{m}\sin\phi \\
	v_y\\
	-\frac{1}{m}\left(mg+C^v_D v_y\right) + \frac{T_1}{ m}\cos\phi + \frac{T_2}{m}\cos\phi\\
	\omega\\
	-\frac{1}{ I_{yy}}C^\phi_D\omega -\frac{ l}{I_{yy}} T_1 + \frac{l}{I_{yy}} T_2
    \end{bmatrix}
\end{align}
\end{subequations} 
with $\underline{T}\le T_1, T_2 \le \bar T$, and where the state is given by the position in the vertical plane $(x,y)$, velocity $(v_x, v_y)$, pitch $\phi$, and pitch rate $\omega$.
The control variables are the thrusts $T_1, T_2$.
In addition, the quadrotor has mass $m$, moment of inertia $I_yy$, half-length $l$.
Furthermore, $g$ denotes the acceleration due to gravity, $C_D^v$ the translational drag coefficient, and $C_D^\phi$ the rotational drag coefficient.

The quadrotor received an observation containing its full state, as well as eight sensor measurements from laser rangefinders. These sensors were located every 45 degrees on the body of the quad, and returned distance to the nearest obstacle. These sensors were fixed to the body frame of the quadrotor. The task of incorporating these sensors outputs to avoid obstacles would require a substantial effort using traditional control, estimation, and mapping tools. 

For efficient computation of approximate BRSs, we decomposed the quadrotor model as follows:

\begin{subequations} 
\label{eq:quad6D_decomp}
\begin{align}
	\begin{bmatrix}
	\dot v_x\\
	\dot\phi
	\end{bmatrix}
	&=
    \begin{bmatrix}
	-\frac{1}{m} C^v_D v_x -\frac{T_1}{m}\sin\phi -\frac{T_2}{ m}\sin\phi \\
	\omega
    \end{bmatrix}, \label{eq:subequations:vxPhi}\\
	\begin{bmatrix}
	\dot v_y\\
	\dot \phi
	\end{bmatrix}
	&=
    \begin{bmatrix}
	-\frac{1}{m}\left(mg+C^v_D v_y\right) + \frac{T_1}{m}\cos\phi + \frac{T_2}{m}\cos\phi \label{eq:subequations:vyPhi}\\
	\omega
    \end{bmatrix}, \\
	\dot x &= 	 v_x, \label{eq:subequations:x}\\
	\dot y &= 	 v_y, \label{eq:subequations:y}\\	
	\dot{ \omega} &= -\frac{1}{ I_{yy}}C^\phi_D\omega -\frac{ l}{ I_{yy}} T_1 + \frac{ l}{ I_{yy}} T_2
\end{align}
\end{subequations} 
with $\underline{T}\le  T_1,  T_2 \le {\bar T}, \underline\omega \le  \omega \le \bar\omega$, $\underline v_x \le  v_x \le \bar v_x$, $\underline v_y \le  v_y \le \bar v_y$, and where $\omega$ is a fictitious control in \eqref{eq:subequations:vxPhi} and \eqref{eq:subequations:vyPhi}, and $v_x$ and $v_y$ are fictitious controls in \eqref{eq:subequations:x} and \eqref{eq:subequations:y} respectively.

\paragraph*{Hyperparameter Settings} In all experiments involving the quadrotor environment, the hyperparameters were set to exactly match the parameters used previously, for the 5-D car environment.

\section{Additional Experimental Results}

\subsection{\expA}

\begin{figure}[t]
\centering
\includegraphics[width=0.9\linewidth]{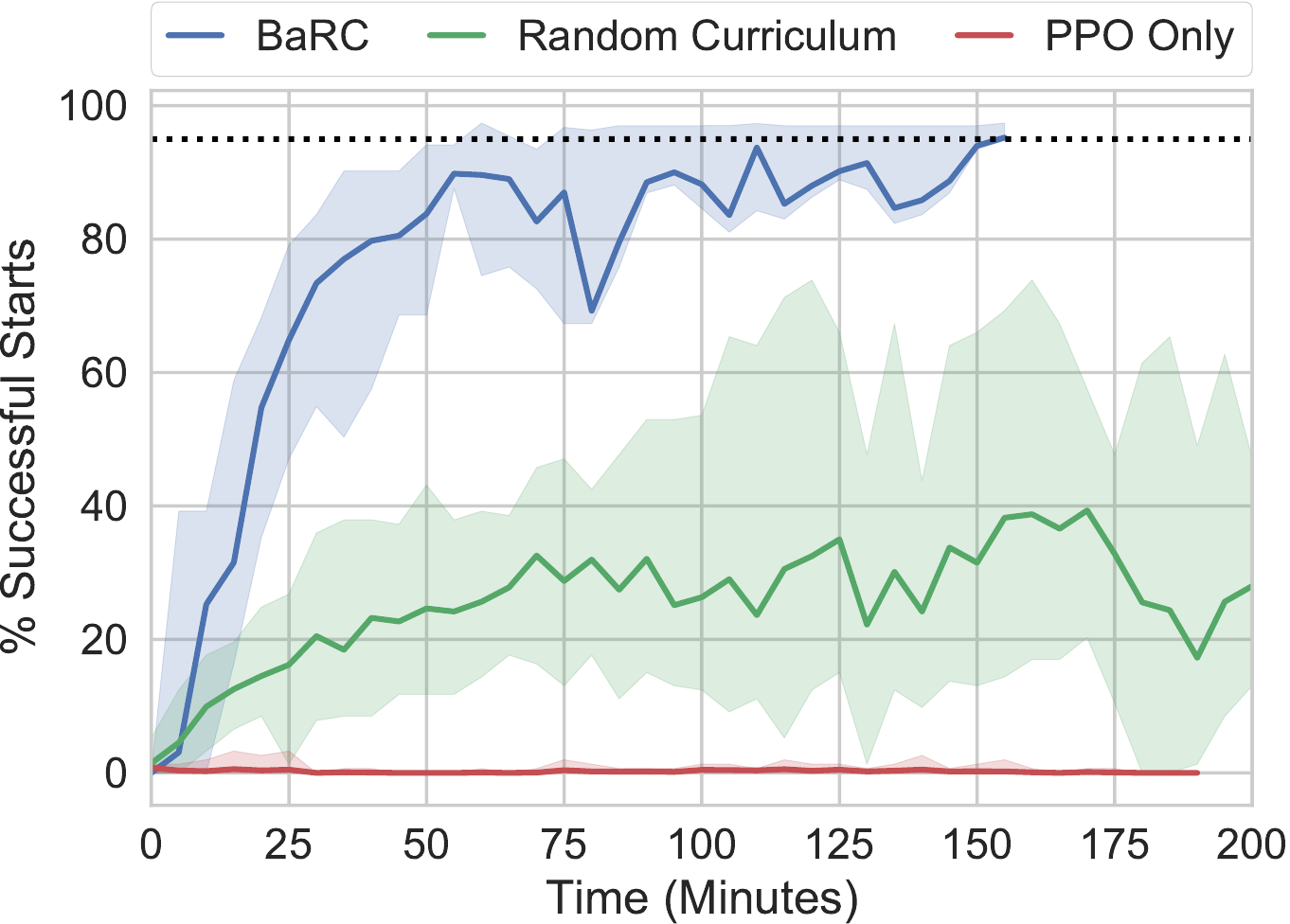}
\caption{Percentage of successful starts for the car model, plotted versus wall clock time.}
\label{fig:car_no_err_minutes}
\end{figure}

\paragraph*{Wall Clock Runtime} In addition to the analyses performed in the main body of the paper, we also compared the wall clock runtime of \algName{} to that of a random curriculum and standard PPO. This experiment was run on servers with octa-core CPUs, but each individual run was limited to running in one thread. Figure \ref{fig:car_no_err_minutes} illustrates our results. We can see that our algorithm is competitive in runtime, even with the additional cost of computing BRSs at each iteration. In this environment, the longest iterations of \algName{} takes 2-3 times as long as standard PPO iterations. However, this multiple rapidly shrinks as the policy learns to reach the goal quickly and learns to generalize across the state space. Once the BRS has reached the initial state, the BRS expansion can be terminated and each iteration of \algName{} runs for approximately the same amount of time as standard PPO.

\paragraph*{Robustness to Model-System Mismatch} To determine how accurate the approximate dynamics model must be (and therefore how robust our approach is to model error), we evaluated \algName{}'s performance in the presence of three common model mismatch scenarios: additive velocity noise, nonzero-mean additive control noise, and oversteer. 

\begin{figure*}[t]
\centering
\includegraphics[width=0.32\linewidth]{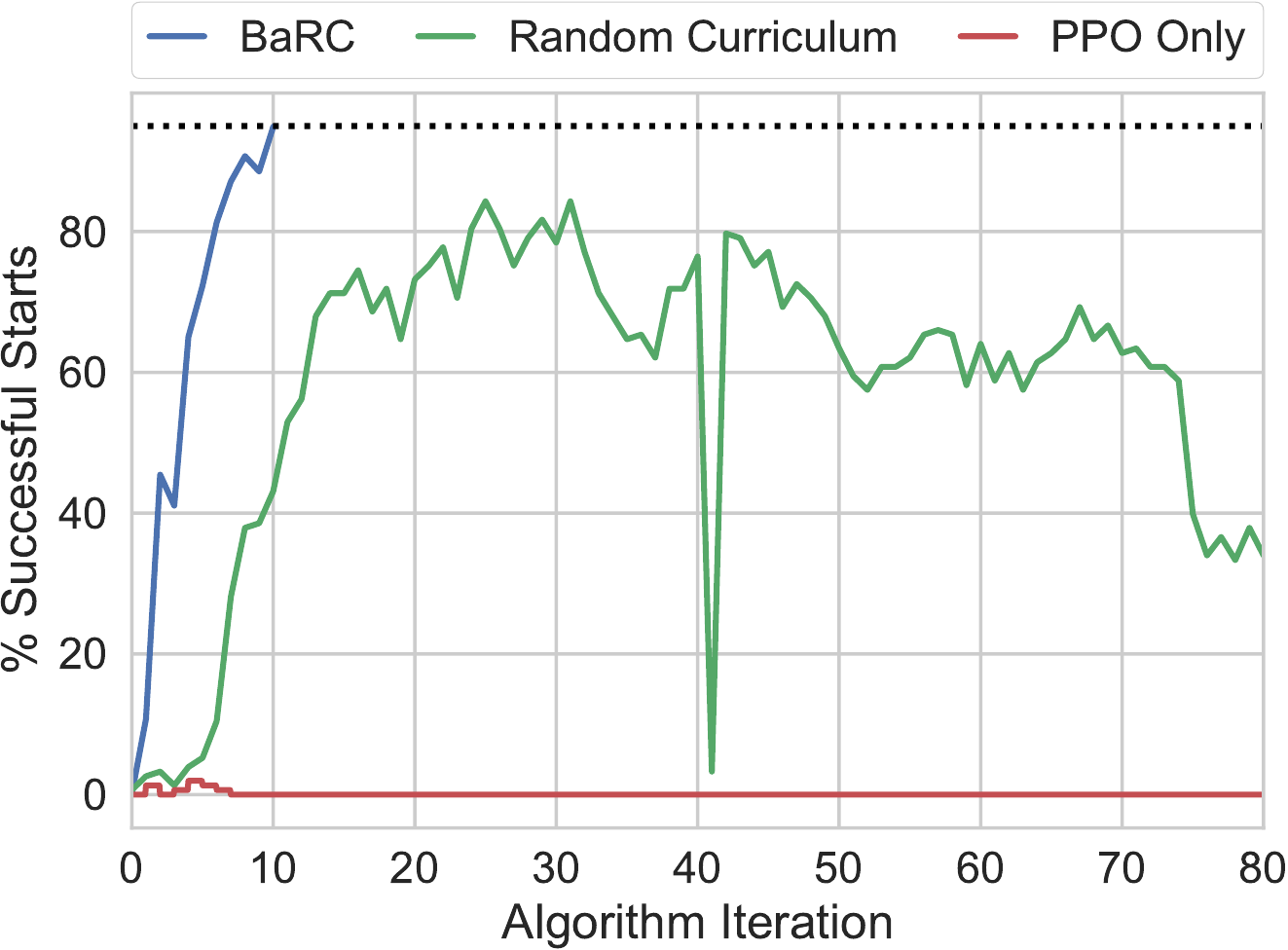}
\includegraphics[width=0.32\linewidth]{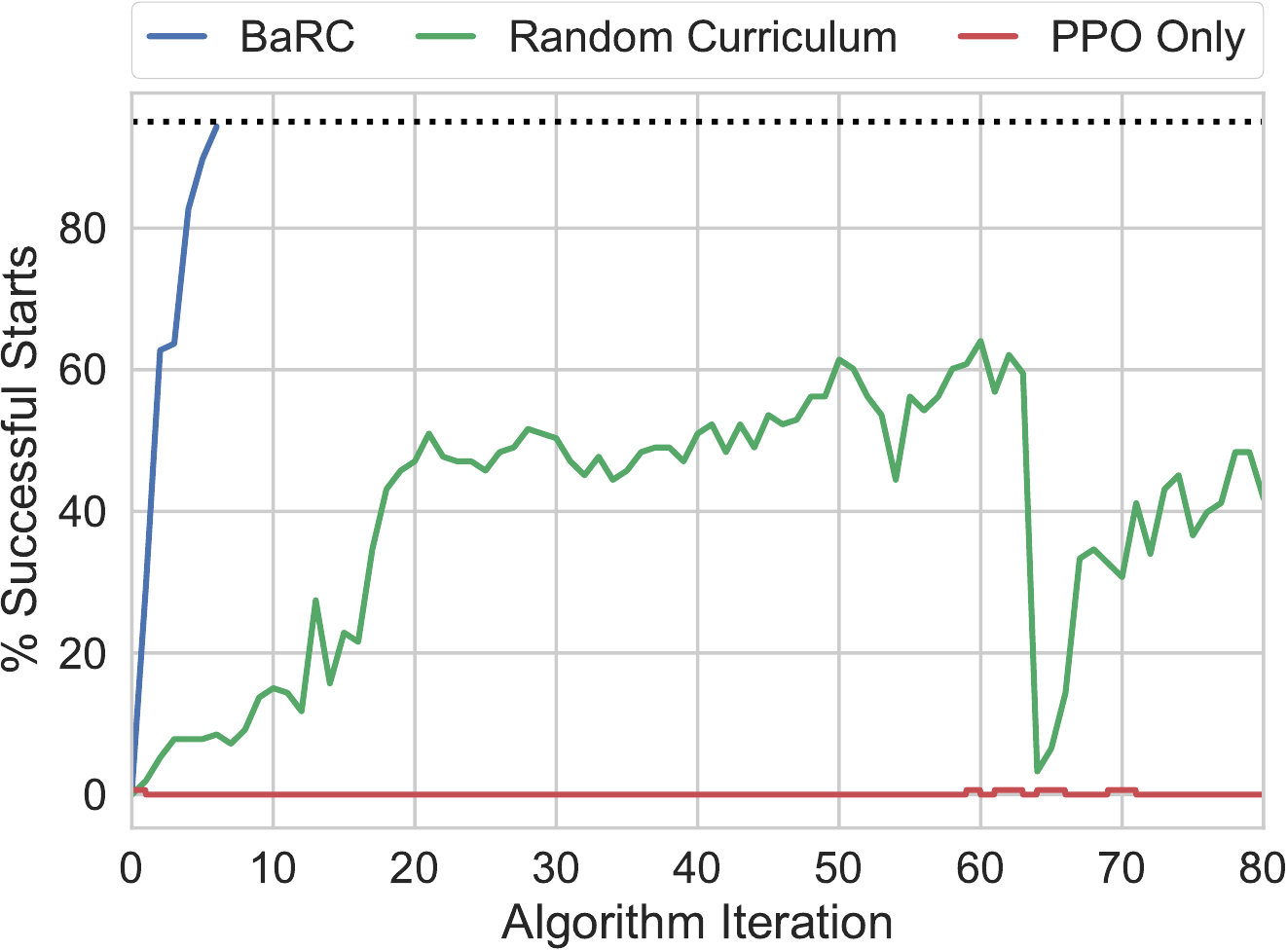}
\includegraphics[width=0.32\linewidth]{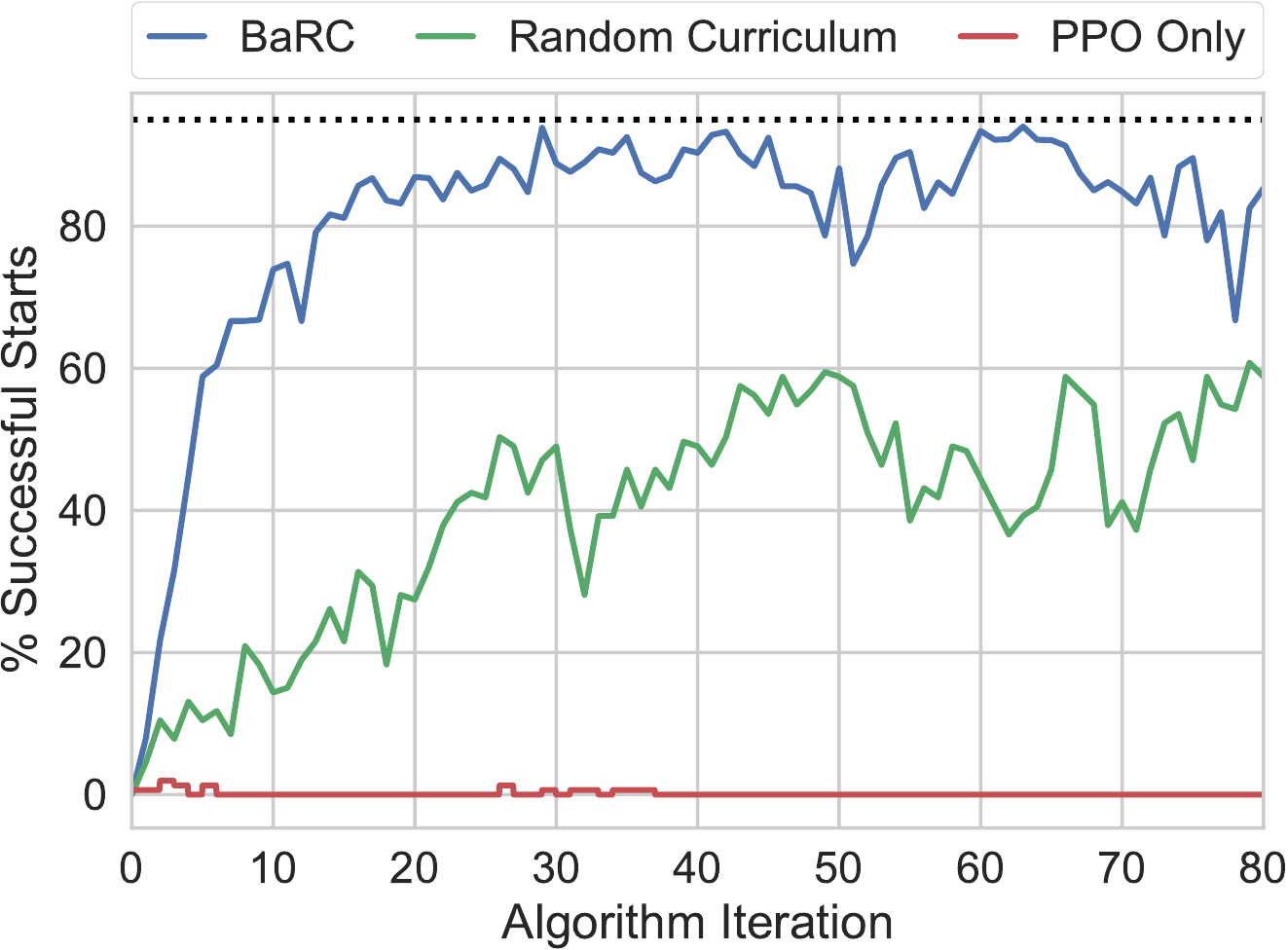}
\caption{Corrupted car model results. \textbf{Left:} The performance of \algName{}, random curricula, and PPO alone on a system with additive nonzero-mean Gaussian noise to the control input. \textbf{Middle:} The performance of the three algorithms on a system with oversteer (steering control inputs are multiplied by a constant factor greater than 1). \textbf{Right:} The algorithms' performance on a system with additive Gaussian longitudinal velocity noise.}
\label{fig:car_err_expts}
\end{figure*}

To add velocity noise, we sample a disturbance value $d_v$ from a standard Gaussian distribution, add this to the current state's velocity, numerically integrate the dynamics equations to produce the next state, and then remove the added velocity from the resulting state. Thus, the added velocity noise affects the propagation of the system, but is not observed by the policy. This velocity noise is roughly equivalent to wheel slip or increased traction during a timestep of the discrete dynamics. For nonzero-mean additive control noise, we sample the disturbance values $d_{v-\text{ctrl}}$ and $d_{\kappa-\text{ctrl}}$ independently from a Gaussian centered at 0.3 with standard deviation 0.2, which are then added to the control inputs. Finally, to produce oversteer we set $d_{a_\kappa} = 1.5$ which multiplies the commanded rate of curvature. Figure \ref{fig:car_err_expts} shows the performance of \algName{}, a random curriculum, and standard PPO on each of these model mismatch scenarios. As can be seen, our approach outperforms the others in all cases. Interestingly, these model mismatches generally improved the performance of the random curriculum method.

\subsection{\expB}

\begin{figure*}[t]
\centering
\includegraphics[width=0.49\linewidth]{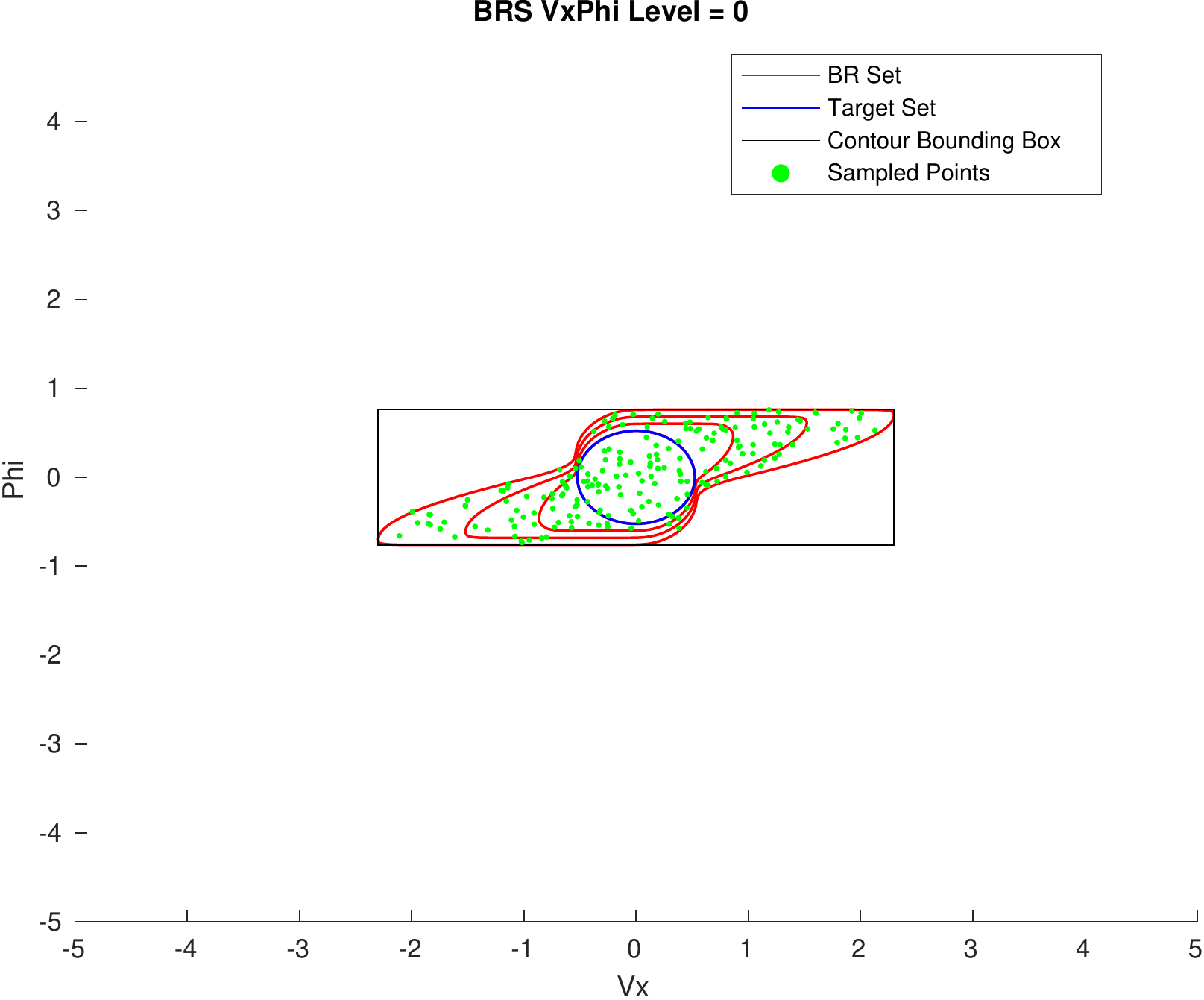}
\includegraphics[width=0.49\linewidth]{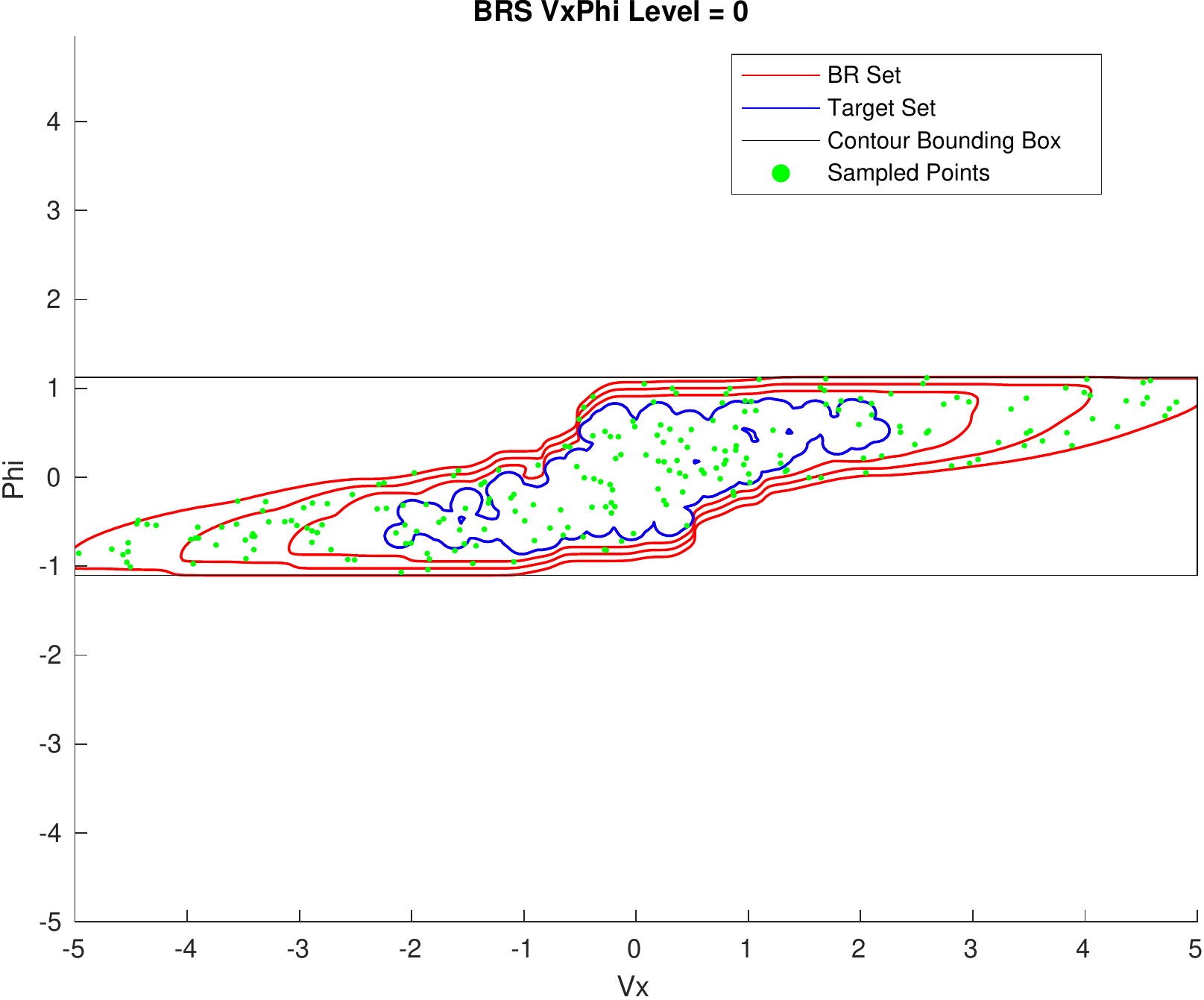}
\includegraphics[width=0.49\linewidth]{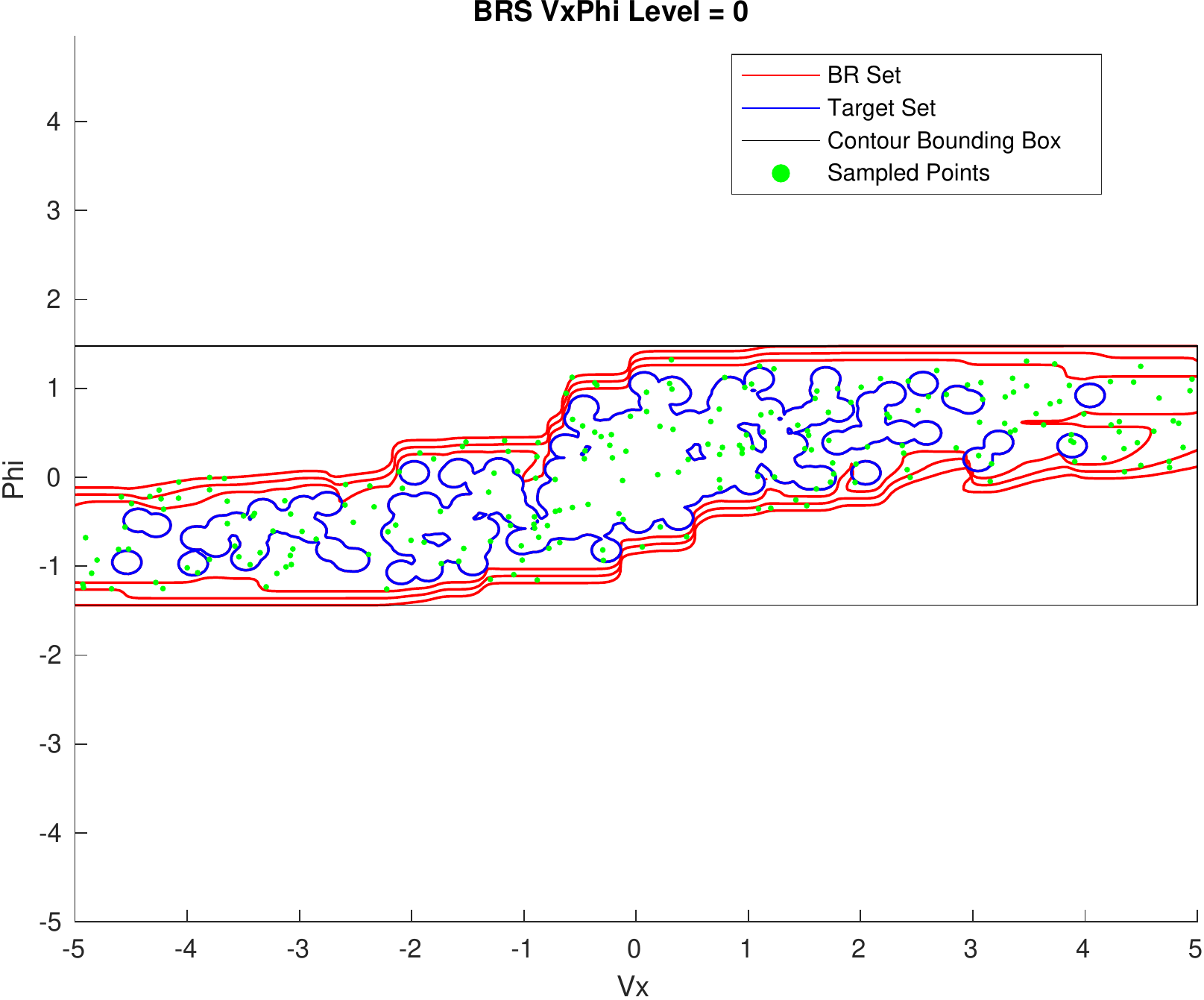}
\includegraphics[width=0.49\linewidth]{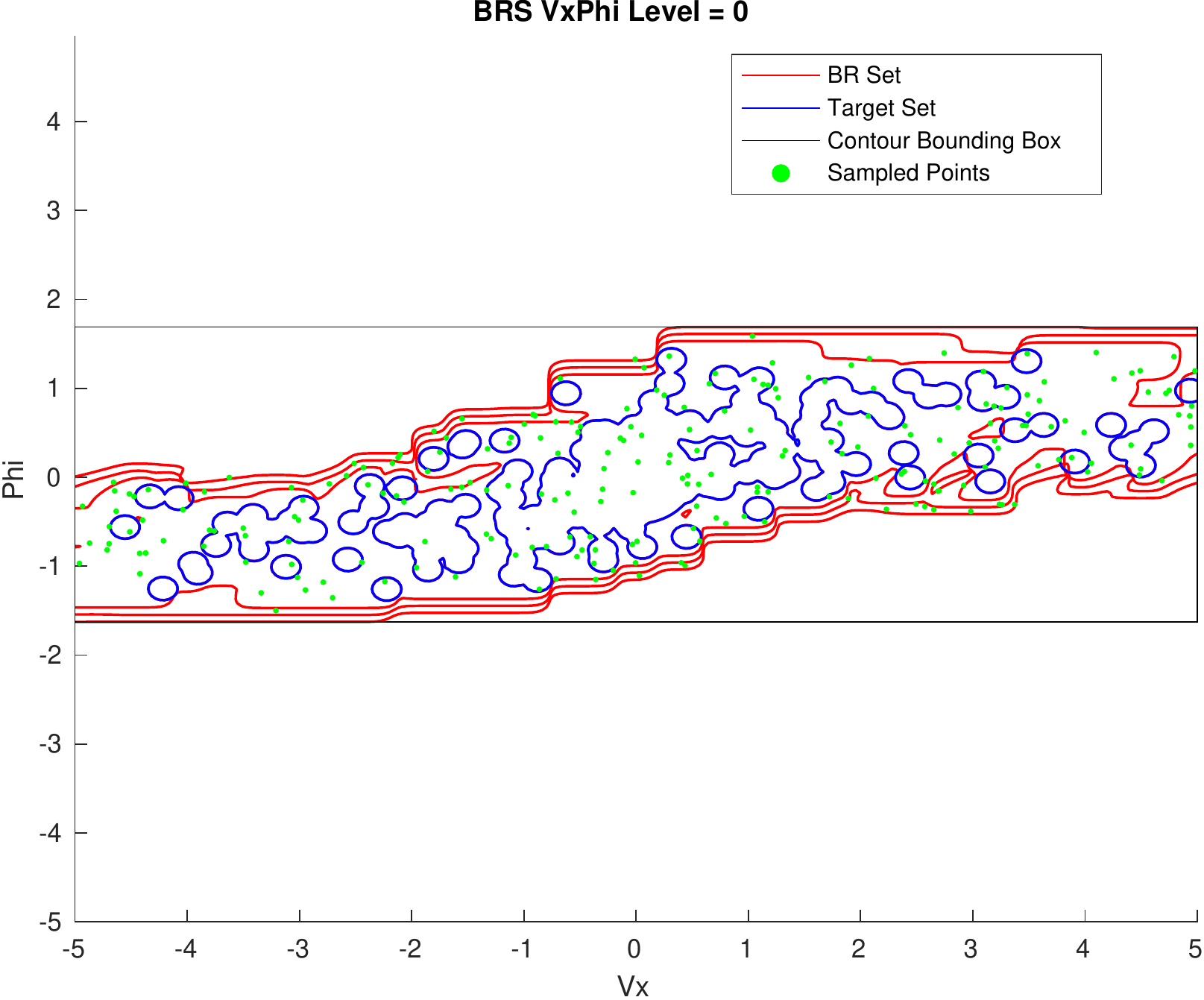}
\caption{An example BRS for our planar quad system and how it evolves over time as more points are added to the starts variable. Plots advance forward from top-left to bottom-right. Green dots show training start states obtained via rejection sampling. The three BRSs shown in red correspond to time horizons of $T = \{0.05, 0.10, 0.15\}$ s, respectively, from the innermost (closest to the target set) to the outermost contour.}
\label{fig:brs_evolution}
\end{figure*}

\paragraph*{BRS Visualization} While not strictly an experiment, for the benefit of the reader we show example BRSs for the planar quadrotor system in Figure \ref{fig:brs_evolution} to illustrate how it is sampled from and what its evolution per iteration looks like.


\end{document}